\documentclass[runningheads]{llncs}
\usepackage{csquotes}

\usepackage[T1]{fontenc}
\usepackage{graphicx}
\usepackage{amssymb}
\usepackage{booktabs}
\usepackage{breakurl}    % for URLs in bib entries (if using biblatex)
\usepackage{url}         % allows line breaks in \url{}
\usepackage{hyperref} % also works with \href{}
\usepackage{subcaption}

\usepackage[english]{babel}
\usepackage{microtype}
\usepackage{graphicx} % for \rotatebox
\usepackage{subcaption}
\usepackage{graphicx} % in preamble

\usepackage{amsmath}
\usepackage[
  backend=biber,
  style=numeric,
  sorting=none
]{biblatex}

\begin{document}
\title{Anatomy-Informed Deep Learning for Abdominal Aortic Aneurysm Segmentation}
\titlerunning{Anatomy-Informed AAA Segmentation}
%
%\titlerunning{Abbreviated paper title}
% If the paper title is too long for the running head, you can set
% an abbreviated paper title here
%
\author{Osamah Sufyan\inst{1}\and
Martin Brückmann\inst{1} \and
Ralph Wickenhöfer\inst{2} \and
Babette Dellen\inst{1} \and
Uwe Jaekel\inst{1}}
\authorrunning{O. Sufyan et al.}
% First names are abbreviated in the running head.
% If there are more than two authors, 'et al.' is used.
%
\institute{Faculty of Mathematics, Informatics and Technology, University of Applied Sciences Koblenz, 53424 Remagen, Germany
\\
 \and
Department of Radiology, Herz-Jesu-Krankenhaus, 56428 Dernbach, Germany\\
\email{sufyan@hs-koblenz.de}}
\maketitle              % typeset the header of the contribution
\vspace{-20pt}  % adjust: -5pt to -15pt

\begin{abstract}
In CT angiography, the accurate segmentation of abdominal aortic aneurysms
(AAAs)  is difficult due to large
anatomical variability, low-contrast vessel boundaries,
and the close proximity of organs whose intensities resemble vascular structures, often leading to false positives. To address these challenges, we propose an
anatomy-aware segmentation framework that integrates
organ exclusion masks derived from TotalSegmentator
into the training process. These masks encode explicit
anatomical priors by identifying non-vascular organs
and penalizing aneurysm predictions within these regions, thereby guiding the U-Net to focus on the aorta
and its pathological dilation while suppressing anatomically implausible predictions. Despite being trained
on a relatively small dataset, the anatomy-aware model
achieves high accuracy, substantially reduces false positives, and improves boundary consistency compared to
a standard U-Net baseline. The results demonstrate that
incorporating anatomical knowledge through exclusion
masks provides an efficient mechanism
to enhance robustness and generalization, enabling reliable AAA segmentation even with limited training
data.
\vspace{-5pt}  % adjust: -5pt to -15pt

\end{abstract}

\section{Introduction}
\vspace{-10pt}  % adjust: -5pt to -15pt

An abdominal aortic aneurysm (AAA) is a life-threatening dilation of the aorta associated with a rupture mortality rate exceeding 80\% \cite{LLoyd2023_AbdominalAorticAneurysm}. Accurate and early assessment is therefore critical for clinical decision-making. Computed tomography angiography (CTA) is the clinical reference standard due to its high spatial resolution and ability to visualize the lumen, thrombus, and surrounding anatomy \cite{Caradu2024_CTA_variability}. However, manual segmentation—particularly of the outer aneurysm wall—is time-consuming and prone to inter-observer variability \cite{Caradu2024_CTA_variability}.

\begin{figure}[htbp]
\centering

\begin{subfigure}[t]{0.38\textwidth}
\vspace{0pt}
\centering
\includegraphics[width=\textwidth]{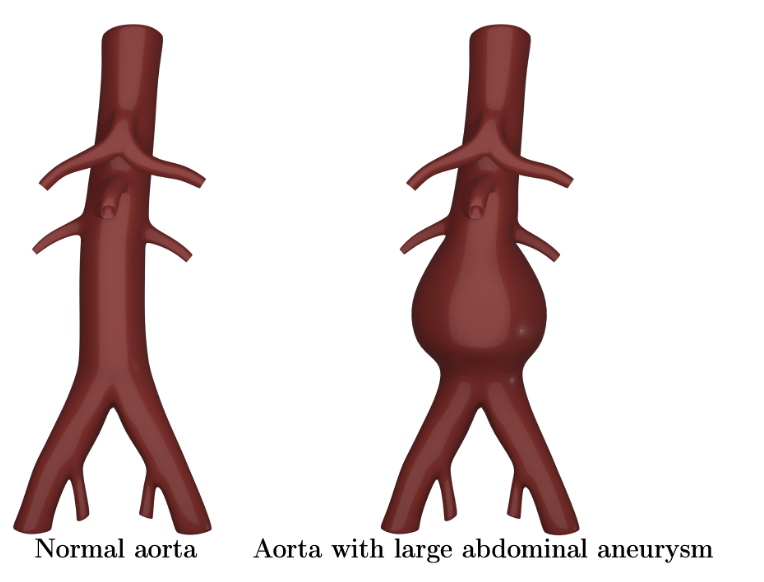}
\caption{}
\label{fig:aaa_anatomy}
\end{subfigure}
\hfill
\begin{subfigure}[t]{0.60\textwidth}
\vspace{0pt}
\centering
\includegraphics[width=\textwidth]{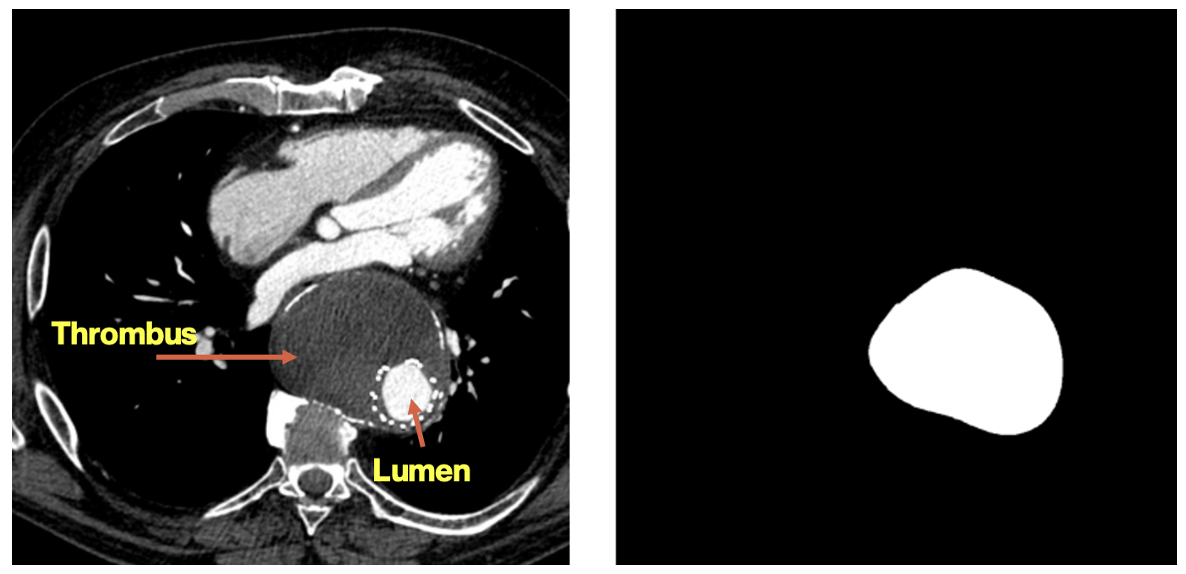}
\caption{}
\label{fig:aaa_ct}
\end{subfigure}

\caption{(a) Comparison between a normal aorta and an aorta with a large aneurysm. (b) Example axial CT slice with lumen, thrombus, and corresponding ground-truth segmentation.}
\label{fig:combined}
\end{figure}

Deep learning, particularly U-Net–based encoder–decoder architectures, has significantly advanced medical image segmentation \cite{Ronneberger2015_UNet} and has been successfully applied to AAA analysis \cite{Sahoo2025_nnUNet_AAA}. However, these methods typically require large annotated datasets, which are scarce in AAA due to the difficulty of labeling the aneurysm wall and challenges such as low contrast, thrombus heterogeneity, and anatomical variability. Segmentation performance is further degraded near anatomically similar structures (e.g., vertebrae, bowel, kidneys), leading to false positives \cite{Fu2021_MultiOrganSeg}. While prior work has incorporated anatomical context through multi-organ segmentation, shape priors, or attention mechanisms, these approaches often increase annotation effort, model complexity, or computational cost \cite{Fu2021_MultiOrganSeg,ShapePrior2024_DeepSeg,Schlemper2018_AttentionGates}.

TotalSegmentator enables automatic segmentation of 104 anatomical structures from CT images without manual labeling \cite{Wasserthal2023_TotalSegmentator}, offering a practical source of anatomical context. In this work, we propose an anatomy-aware AAA segmentation framework that integrates organ exclusion masks derived from TotalSegmentator into U-Net training. These masks identify non-vascular regions and penalize anatomically implausible predictions, reducing false positives and improving boundary consistency without requiring architectural changes or additional annotations. Despite being trained on a relatively small dataset, the proposed method achieves high accuracy and improved robustness compared to a baseline U-Net. To our knowledge, this is the first work to leverage automatically generated multi-organ masks as exclusion priors for AAA segmentation.

\section{Methods}
\vspace{-5pt}  % adjust: -5pt to -15pt

\subsection{Dataset}

The dataset consists of abdominal CT angiography (CTA) scans from 20 patients with confirmed abdominal aortic aneurysms, annotated under expert supervision by a senior radiologist. Each scan covers the abdominal aorta from the diaphragmatic hiatus to the iliac bifurcation, with slice thickness between 0.6 and 1.0 mm. All data were anonymized and stored in DICOM format.

Scans were converted to NIfTI using dcm2niix and resampled to an isotropic resolution of $1.0 \times 1.0 \times 1.0 \, mm^3$. Axial slices were extracted and resized to $512 \times 512$, with intensities normalized to [0, 255] after applying a soft-tissue window (–150 to 600 HU). Slices not containing the abdominal aorta were automatically excluded using TotalSegmentator-based anatomical filtering.

Manual ground-truth masks delineating the outer aneurysm wall, including thrombus, were generated for all relevant slices. The dataset was partitioned at the patient level to prevent slice-level data leakage. Specifically, 10 patients were used for training, while 2 patients were reserved for validation. An additional 3 patients were allocated for testing. Each patient contributed approximately 150–300 slices. In scenarios where 16 patients were available for training, 4 patients were instead assigned to the test set.

To improve generalization given the limited dataset size, extensive on-the-fly data augmentation was applied, including rotations (±10°), flips, translations (±10 pixels), scaling (0.9–1.1×), intensity variations, Gaussian noise, and elastic deformations. All preprocessing and augmentation steps were implemented in Python using PyTorch-based pipelines.
\vspace{-10pt}  % adjust: -5pt to -15pt

\begin{figure}[ht]
\centering
\includegraphics[width=0.6\linewidth]{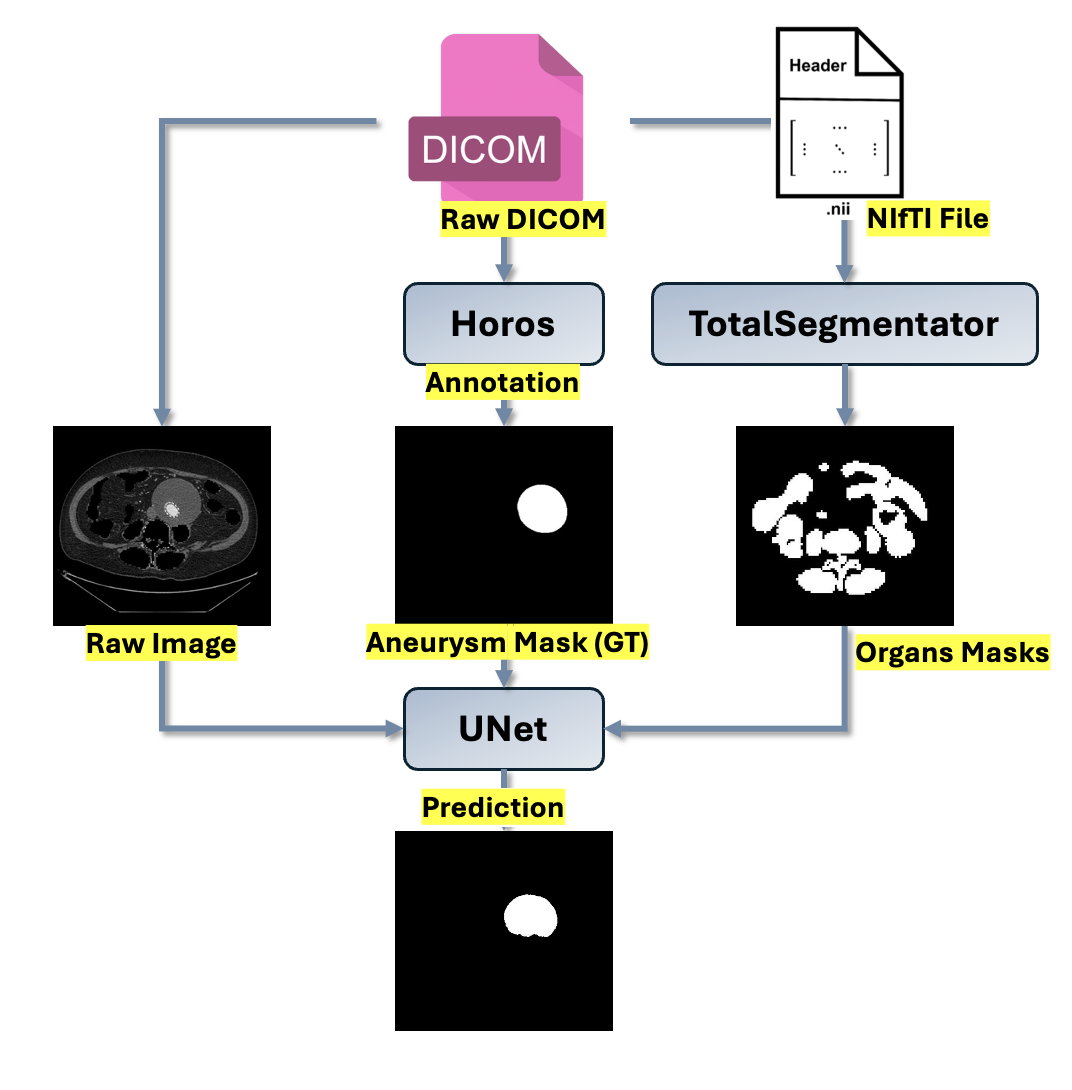}
\caption{Pipeline of the proposed method from preprocessing to training.}
\label{fig:false-color}
\end{figure}

\vspace{-30pt}  % adjust: -5pt to -15pt

\subsection{Anatomy-Aware Organ Exclusion and Masked Loss}
\vspace{-5pt}

To incorporate anatomical priors, we applied \textit{TotalSegmentator}~\cite{Wasserthal2023_TotalSegmentator}, a 3D U-Net–based framework that automatically segments 104 anatomical structures from CT volumes. From these outputs, non-vascular abdominal structures were merged into a binary \textit{organ exclusion mask}, while the aorta and iliac arteries were excluded to preserve relevant vascular regions. For each axial slice \(X\), the exclusion mask was resampled and aligned with the corresponding aneurysm ground truth \(Y\).

This exclusion mask is converted into a binary allow mask \(A\), where \(A_i=1\) denotes anatomically plausible regions and \(A_i=0\) excluded regions. Let \(P, Y, A \in \mathbb{R}^N\) denote flattened prediction, ground truth, and mask (\(N=H\cdot W\)). Loss computation is restricted to pixels with \(A_i=1\).

The masked Dice and binary cross-entropy (BCE) losses are defined as
\begin{equation}
\mathcal{L}_{\text{Dice}} = 1 - 
\frac{2\, (A \cdot P \cdot Y) + \epsilon}
     {(A \cdot P) + (A \cdot Y) + \epsilon},
\end{equation}
\begin{equation}
\mathcal{L}_{\text{BCE}} =
\frac{A \cdot \mathrm{BCE}(P, Y)}
     {A \cdot \mathbf{1} + \epsilon},
\end{equation}
where \(\cdot\) denotes element-wise multiplication followed by summation. The total loss is
\begin{equation}
\mathcal{L} = w\,\mathcal{L}_{\text{BCE}} + (1-w)\,\mathcal{L}_{\text{Dice}}, \quad w=0.5.
\end{equation}

This formulation removes anatomically implausible regions from optimization rather than penalizing them, reducing noisy gradients and improving generalization without requiring architectural changes or additional annotations.
\vspace{-10pt}  % adjust: -5pt to -15pt

\subsection{Network Architecture}
\vspace{-5pt}

We employ a two-dimensional U-Net for aneurysm segmentation, motivated by the limited dataset size, where 2D models are less prone to overfitting and allow larger batch sizes and stronger data augmentation. 

The network follows a standard encoder–decoder design with four resolution levels. Each encoder block consists of two \(3 \times 3\) convolutions with ReLU and batch normalization, followed by \(2 \times 2\) max-pooling. Feature channels increase from 32 to 256. The decoder mirrors this structure using transposed convolutions for upsampling and incorporates skip connections to preserve fine anatomical details.

A final \(1 \times 1\) convolution with sigmoid activation produces a single-channel probability map. No architectural modifications were introduced, allowing isolation of the effect of anatomical priors.

Training was performed end-to-end using the Adam optimizer (\(10^{-4}\)), a batch size of 3, and 200--300 epochs with early stopping based on validation Dice. Implementation was done in PyTorch on an NVIDIA L40S GPU.

\vspace{-10pt}  % adjust: -5pt to -15pt

\subsection{Baseline Models}
\vspace{-2pt}  % adjust: -5pt to -15pt

To evaluate the proposed anatomy-aware approach, we compare it to a standard U-Net trained without anatomical priors. The baseline uses the same preprocessing, training schedule, optimizer, and data augmentation, ensuring that performance differences arise solely from the incorporation of organ exclusion masks.

The model is trained with a conventional Dice loss without spatial masking, such that all pixels contribute equally to the optimization. As a result, aneurysm segmentation must be learned purely from image intensity and contextual information.

For a fair comparison, both models are initialized consistently and trained under identical conditions, with early stopping based on the validation Dice score. This setup allows improvements to be attributed directly to the anatomy-aware formulation.

\section{Results}
\subsection{Quantitative Evaluation}
The proposed anatomy-aware model demonstrated consistently higher segmentation accuracy than the standard U-Net baseline for all test patients. Table 1 summarizes the slice-wise Dice coefficients for both models. For each patient, we report the mean Dice score, the standard deviation for all slices, and the number of slices in which aneurysm tissue was present.

\begin{figure}
\centering

% -------- Left --------

\begin{subfigure}[t]{0.45\textwidth}
    \vspace{0pt}
    \centering

    \newcommand{\w}{0.42\linewidth}
    \newcommand{\rw}{0.12\linewidth}

    % Row 1
    \begin{minipage}{\rw}
        \centering
        \rotatebox{90}{\textbf{Patient 1}}
    \end{minipage}
    \begin{minipage}{\w}
        \includegraphics[width=\linewidth]{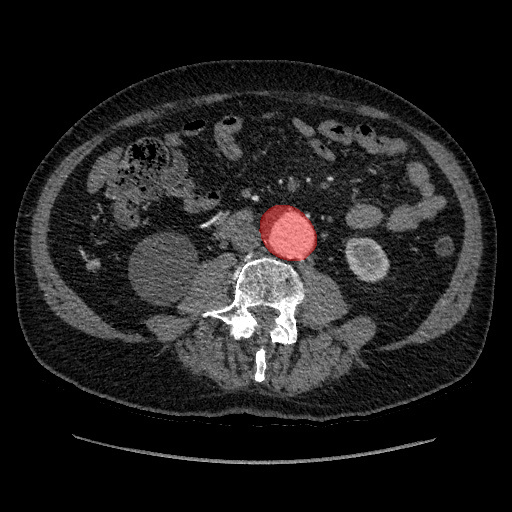}
    \end{minipage}
    \begin{minipage}{\w}
        \includegraphics[width=\linewidth]{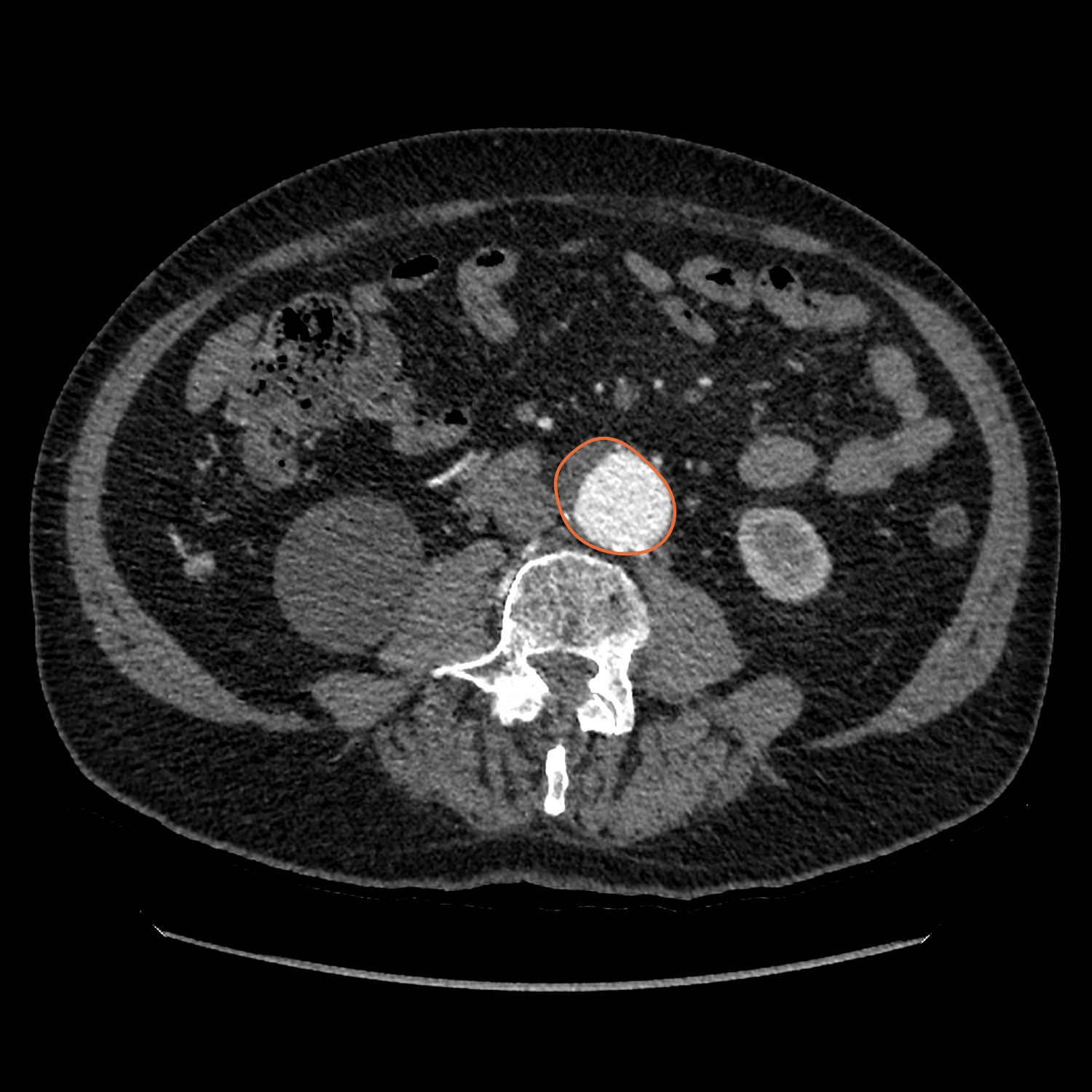}
    \end{minipage}

    \vspace{3mm}

    % Row 2
    \begin{minipage}{\rw}
        \centering
        \rotatebox{90}{\textbf{Patient 2}}
    \end{minipage}
    \begin{minipage}{\w}
        \includegraphics[width=\linewidth]{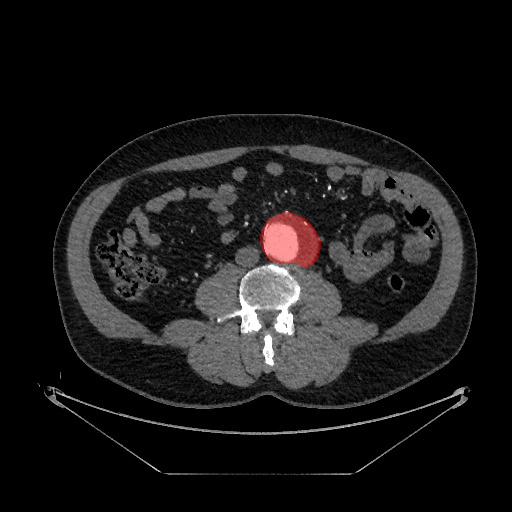}
    \end{minipage}
    \begin{minipage}{\w}
        \includegraphics[width=\linewidth]{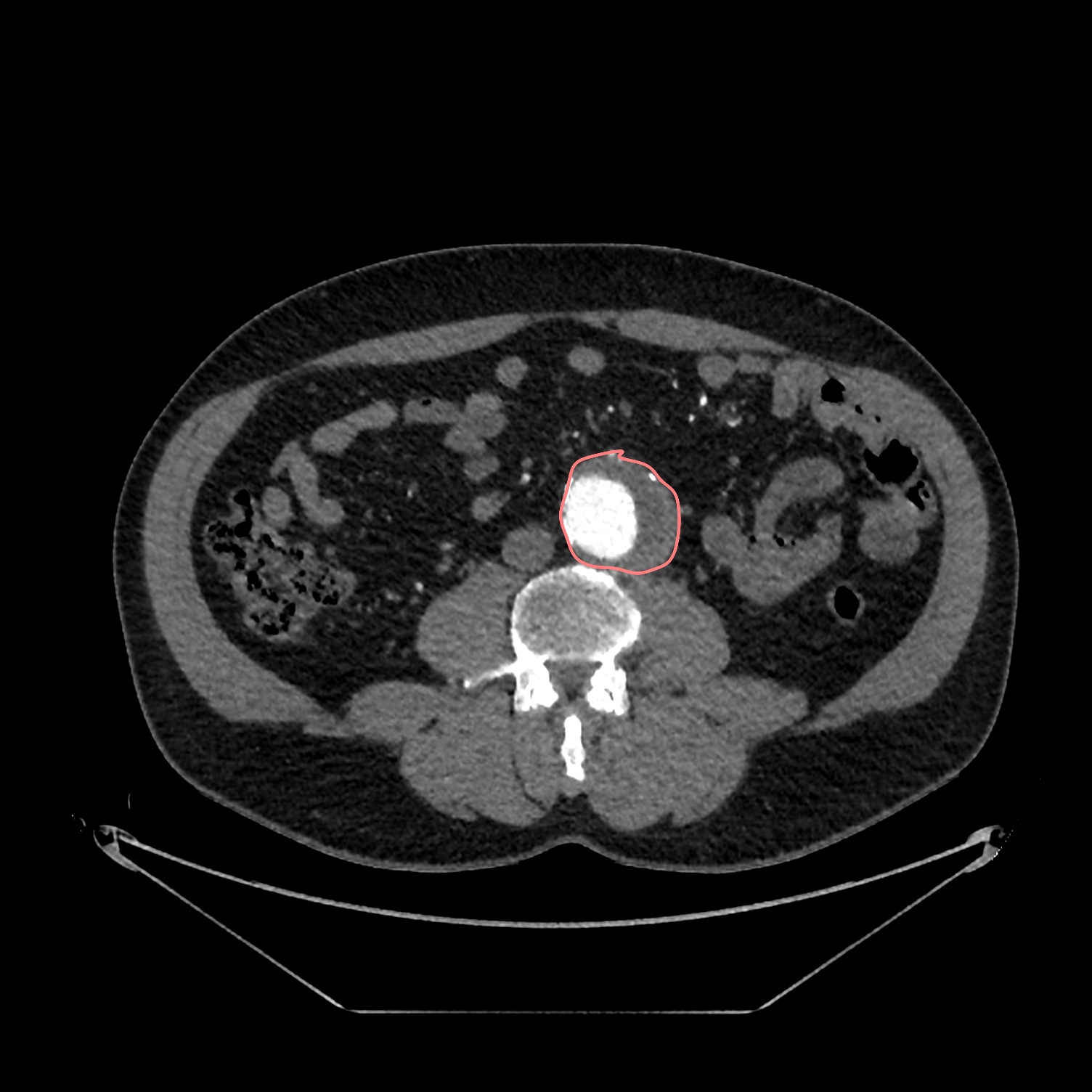}
    \end{minipage}

    \vspace{3mm}

\hspace{\rw}
\makebox[\w][c]{\fontsize{4pt}{10pt}\selectfont\textbf{Model segmentation}}
\makebox[\w][c]{\fontsize{4pt}{10pt}\selectfont\textbf{Ground truth}}

    \caption{}
\end{subfigure}
\hfill
\begin{subfigure}[t]{0.50\textwidth}
    \vspace{1.4cm} % <-- adjust this value
    \centering
    \includegraphics[width=\linewidth]{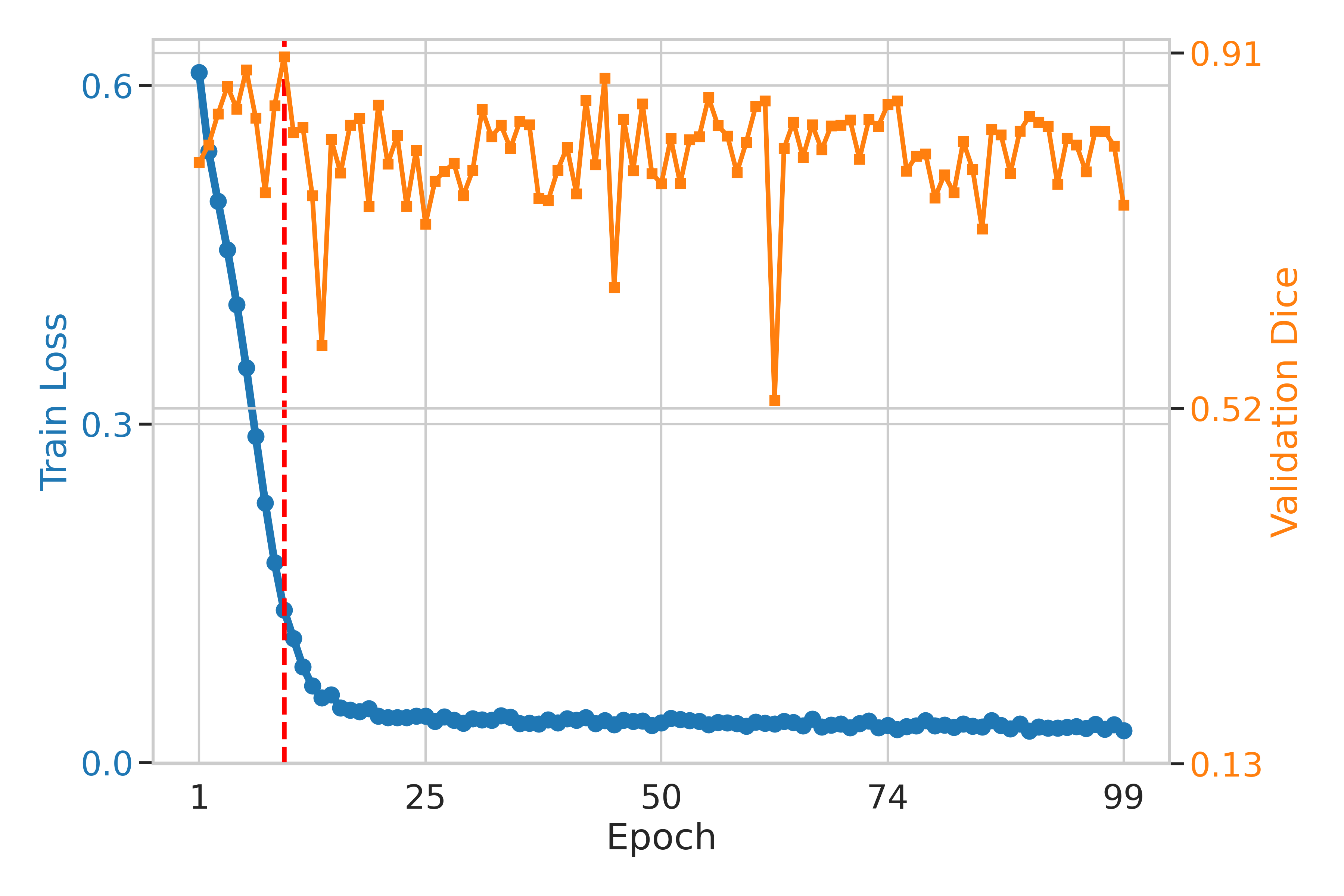}
    \caption{}
\end{subfigure}

% -------- Right --------

\caption{(a) Qualitative comparison showing two representative slices from different
patients. Each row corresponds to a different patient case. The model correctly segments the aneurysm with thrombus. (b) Loss curve and validation DICE vs epoch with the epoch with the best
validation result marked with vertical red line }
\end{figure}
\vspace{-40pt}  % adjust: -5pt to -15pt

\begin{table}[htbp]
\centering
\captionsetup{font=footnotesize}
\caption{Model comparison on the test set.}
\label{tab:global_results}

\setlength{\tabcolsep}{3.5pt}
\renewcommand{\arraystretch}{1.1}
\footnotesize

\begin{tabular}{|l|c|c|c|c|}
\hline
\textbf{Model} & \textbf{Train Patients} & \textbf{Mean Dice} & \textbf{Precision} & \textbf{Recall} \\
\hline
Baseline U-Net & 10 & $0.637 \pm 0.442$ & 0.796 & 0.762 \\
\hline
Anatomy-Aware U-Net & 10 & $\mathbf{0.914 \pm 0.033}$ & \textbf{0.965} & \textbf{0.882} \\
\hline
Anatomy-Aware U-Net & 16 & $\mathbf{0.906 \pm 0.179}$ & \textbf{0.961} & \textbf{0.942} \\
\hline
\end{tabular}
\end{table}

\vspace{-20pt}  % adjust: -5pt to -15pt

Overall, our method achieved higher Dice scores on nearly all patients, with the largest improvements observed in anatomically challenging cases where the aneurysm was adjacent to bowel loops, vertebrae, or renal structures. These gains reflect the utility of organ exclusion masks in reducing false activations in regions with intensity profiles similar to aneurysm thrombus.

The proposed anatomy-aware U-Net consistently outperformed the baseline U-Net across different training-set sizes. Table~\ref{tab:global_results} summarizes the quantitative results. With a batch size of 3 and 10 training patients, the baseline U-Net achieved a mean Dice score of 0.637, while the anatomy-aware U-Net achieved a score of 0.914. Increasing the batch size and training patients further kept the boosted performance, reaching a mean Dice of 0.906 for the anatomy-aware U-Net with 16 training patients and a better recall of 0.942. These results demonstrate that incorporating anatomical priors significantly enhances segmentation accuracy, particularly when training data is limited.

In addition to numerical gains, the anatomy-aware model produced more anatomically plausible segmentations. In Fig.~5, representative examples of the results obtained with the baseline and the proposed model are shown. The baseline U-Net frequently produced false-positive activations in non-vascular structures such as bowel loops, vertebrae, and paraspinal musculature.

Together, the quantitative and visual results demonstrate that the anatomy-aware training strategy improves segmentation quality, reduces anatomically implausible predictions, and enhances robustness, particularly in scenarios with limited training data.
\vspace{-10pt}  % adjust: -5pt to -15pt

\subsection{Three-Dimensional Reconstruction and Morphological Analysis}

We present a pipeline for reconstructing patient-specific 3D vascular geometries and extracting centerline-based morphological descriptors from segmentation masks. Three-dimensional (3D) models were reconstructed from two-dimensional (2D) binary segmentation masks of the aneurysm lumen by aligning slices using imaging metadata and stacking them into a volumetric grid. To reduce voxelization artifacts, the volume was smoothed and converted into a watertight triangular mesh using the marching cubes algorithm, followed by Laplacian and Taubin smoothing to preserve anatomical fidelity (Fig.~\ref{fig:false-color}).

\begin{table}[ht]
\centering
\caption{Predicted and ground-truth geometric measurements for each patient.}
\label{tab:geometry_comparison_transposed}

\resizebox{0.61\linewidth}{!}{
\begin{tabular}{|l|c|c|}
\hline
\textbf{Measurement} & \textbf{Test Patient 1} & \textbf{Test Patient 2} \\
\hline
Maximal Diameter (pred) & 5.3 cm & 4.7 cm \\
\hline
Maximal Diameter (true) & 5.8 cm & 5.0 cm \\
\hline
Surface Area (pred) & 616 cm$^{2}$ & 253 cm$^{2}$ \\
\hline
Surface Area (true) & 647 cm$^{2}$ & 191 cm$^{2}$ \\
\hline
Volume (pred) & 492 cm$^{3}$ & 203 cm$^{3}$ \\
\hline
Volume (true) & 538 cm$^{3}$ & 163 cm$^{3}$ \\
\hline
\end{tabular}
}
\end{table}
\vspace{-10pt}  % adjust: -5pt to -15pt

\begin{figure}[t]
    \centering
    \resizebox{1.0\textwidth}{!}{
    \begin{tabular}{cc}
        \includegraphics[width=0.3\textwidth]{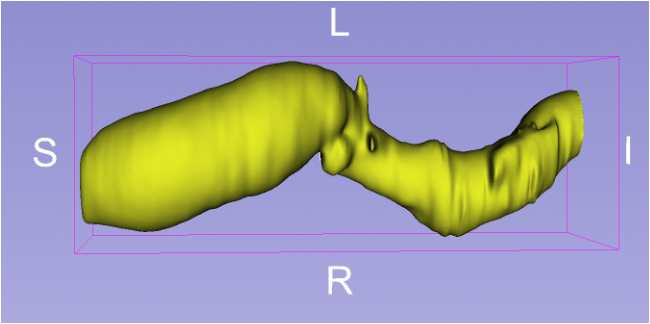} & 
        \includegraphics[width=0.3\textwidth]{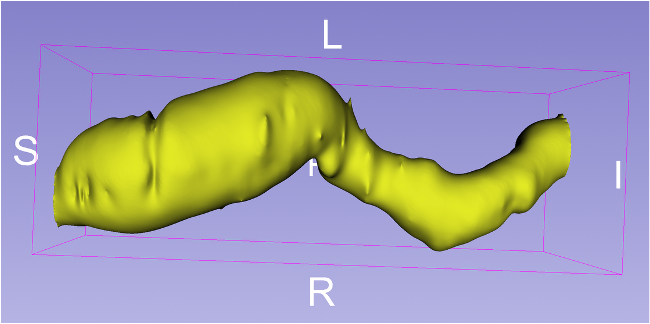} \\
        {\fontsize{6}{6}\selectfont (a) Ground truth} &
        {\fontsize{6}{6}\selectfont (b) Prediction}
    \end{tabular}
    }
    \caption{3D reconstructed aneurysm from predicted segmentation (right) and corresponding ground truth (left).}
    \label{fig:false-color}
\end{figure}

Vascular centerlines were extracted using the Vascular Modelling Toolkit~\cite{VMTK}. This method employs a distance transform to assign each voxel its minimum distance to the vessel wall and defines a traversal cost inversely proportional to this distance. A fast-marching minimal-cost path between inlet and outlet points yields a smooth centerline approximating the medial axis. Local tangent vectors were computed along the centerline, and orthogonal planes were defined at each point to align cross-sections with the local flow direction. Intersecting these planes with the surface mesh produced closed lumen contours, from which anatomical radii and diameters were derived (e.g., maximal inscribed radius, equivalent-area radius, and maximum chord diameter), enabling accurate characterization of asymmetric aneurysm geometries.

These descriptors support computation of clinically relevant metrics such as maximum diameter, surface area, volume, curvature, and shape indices (e.g., aspect and size ratios). Local diameter variations relate to wall stress via Laplace’s law, with enlarged regions experiencing higher mechanical load, and prior studies have linked geometric features—including diameter, curvature, and tortuosity—to aneurysm rupture risk~\cite{Raut2013_GeometricBiomechanics,Teng2022_CurvatureWSS}.

\vspace{-5pt}  % adjust: -5pt to -15pt

\section{Discussion}
\vspace{-5pt}  % adjust: -5pt to -15pt

The results demonstrate that incorporating anatomical priors via organ exclusion masks improves both the robustness and accuracy of abdominal aortic aneurysm segmentation. While the baseline U-Net performed well in high-contrast regions, it frequently produced false positives in adjacent structures such as bowel, vertebrae, and paraspinal muscles. The anatomy-aware model mitigated these errors by penalizing anatomically implausible predictions, resulting in more consistent and realistic segmentations. It also showed strong performance under limited training data, where the baseline degraded, indicating that anatomical priors act as an effective regularization mechanism that enhances generalization in challenging clinical scenarios.

The proposed method is simple and easily integrable, requiring no architectural modifications or additional annotations beyond aneurysm labels. However, its slice-wise design limits the use of full 3D spatial context, which could be addressed in future work through volumetric models or soft anatomical constraints. Overall, this anatomy-aware approach offers a practical and clinically relevant improvement for AAA segmentation.

\paragraph*{\textbf{Acknowledgments}}
This research has received funding from the Ministry of Science
and Health of Rhineland-Palatinate, Germany, and the Debeka Krankenversicherungsverein
a.G. through the Forschungskolleg Data2Health.
\vspace{-10pt}  % adjust: -5pt to -15pt

\paragraph*{\textbf{Informed Consent}}
All patients provided informed consent for the clinically indicated CT examination in accordance with standard clinical practice. Institutional Review Board (IRB) approval was obtained.
\vspace{-10pt}  % adjust: -5pt to -15pt

%
% ---- Bibliography ----
%
% BibTeX users should specify bibliography style 'splncs04'.
% References will then be sorted and formatted in the correct style.
%
% \bibliographystyle{splncs04}
% \bibliography{mybibliography}
%
\printbibliography
         % name of your .bib file without extension

\end{document}